\documentclass[sigconf,natbib=true,anonymous=false]{acmart}

\makeatletter
\renewcommand\@formatdoi[1]{\ignorespaces}
\makeatother

\usepackage[utf8]{inputenc}
\usepackage{algorithmic}

\usepackage[linesnumbered,lined,boxed]{algorithm2e}
\usepackage{verbatim}
\usepackage{xcolor}
\usepackage{float}
\usepackage{tabularx} 
\usepackage{booktabs}
\usepackage{multirow}
\usepackage{mathtools}
\usepackage{setspace}
\usepackage{placeins}

\usepackage{tikz}
\usetikzlibrary{arrows.meta}
\usepackage{pgfplots}
\usepackage{caption}
\usepackage{subfig}

\usepackage{amsmath}
\usepackage{algorithmic}

\usepackage{mathtools}
\usepackage{setspace}
\usepackage{placeins}
\usepackage{caption}
\usepackage{subfig}
\usepackage{pgfplots}
\usetikzlibrary{matrix}
\usepgfplotslibrary{groupplots}

\usepackage{paralist}
\usepackage[shortlabels]{enumitem}

\usepackage{xcolor,pifont}

\newtheorem{definition}{Definition}

\usepgfplotslibrary{statistics}

\usepackage{xcolor,pifont}

\definecolor{cycle2}{RGB}{106, 191, 0}
\definecolor{cycle3}{RGB}{191, 0, 0}

\definecolor{amber}{rgb}{1.0, 0.75, 0.0}
\definecolor{awesome}{rgb}{1.0, 0.13, 0.32}
\definecolor{ao(english)}{rgb}{0.0, 0.5, 0.0}

\newcommand{\tuple}[1]{\ensuremath{\left \langle #1 \right \rangle }}

\title{OntoMerger: An Ontology Integration Library for 
Deduplicating and Connecting Knowledge Graph Nodes}

\author{David Geleta}
\affiliation{
  \institution{AstraZeneca}
  \country{United Kingdom}
}

\author{Andriy Nikolov}
\affiliation{
  \institution{AstraZeneca}
  \country{United Kingdom}
}

\author{Mark O'Donoghue}
\affiliation{
  \institution{AstraZeneca}
  \country{United Kingdom}
}

\author{Benedek Rozemberczki}
\affiliation{
  \institution{AstraZeneca}
  \country{United Kingdom}
}

\author{Anna Gogleva}
\affiliation{
  \institution{AstraZeneca}
  \country{United Kingdom}
}

\author{Valentina Tamma}
\affiliation{
  \institution{The University of Liverpool}
  \country{United Kingdom}
}

\author{Terry R. Payne}
\affiliation{
  \institution{The University of Liverpool}
  \country{United Kingdom}
}

\begin{document}

\begin{abstract}
Duplication of nodes\footnote{For example, the concept of the disease \textit{Asthma} may be represented as $\mathsf{MONDO:0004979}$ or $\mathsf{DOID:2841}$ in two different datasets.} is a common problem encountered when building knowledge graphs (KGs) from heterogeneous datasets, where it is crucial to be able to merge nodes having the same meaning. OntoMerger is a Python ontology integration library whose functionality is to deduplicate KG nodes. Our approach takes a set of KG nodes, mappings and disconnected hierarchies and generates a set of merged nodes together with a connected hierarchy. In addition, the library provides analytic and data testing functionalities that can be used to fine-tune the inputs, further reducing duplication, and to increase connectivity of the output graph. OntoMerger can be applied to a wide variety of ontologies and KGs. In this paper we introduce OntoMerger and illustrate its functionality on a real-world biomedical KG.
\end{abstract}
\maketitle

\section{Introduction}
Complex knowledge graphs (KGs) \cite{geleta2021biological, sinha2015overview} are often built from multiple overlapping but heterogeneous data sets. For example, a biomedical KG can include separate data sets providing information about genes, diseases, compounds and relations between these entities \cite{geleta2021biological,himmelstein2017systematic,biosnapnets,breit2020openbiolink,walsh2020biokg,10.1007/978-3-030-00668-6_18}. Typically, these data sets rely on diverse ontologies to refer to the same concepts with no obvious mapping or alignment between them \cite{bonner2021review}. Furthermore, data version discrepancy can also lead to duplication when concepts get deprecated and identifiers reassigned. On a large scale, these issues may cause syntactic and semantic heterogeneity in the resulting graph; which in turn can lead to fragmented knowledge that can negatively impact downstream 
applications \cite{hamilton2020graph} due to possible incompatibilities between the hierarchies, that in turn may lead to logical inconsistencies.
Thus, organising concepts of the same domain into a hierarchy \cite{wang2021leveraging} might improve the accuracy of machine learning models, e.g. pattern recognition.

\begin{figure}[!ht]
\centering
\resizebox{.80\linewidth}{!}{
    \includegraphics[width=\linewidth]{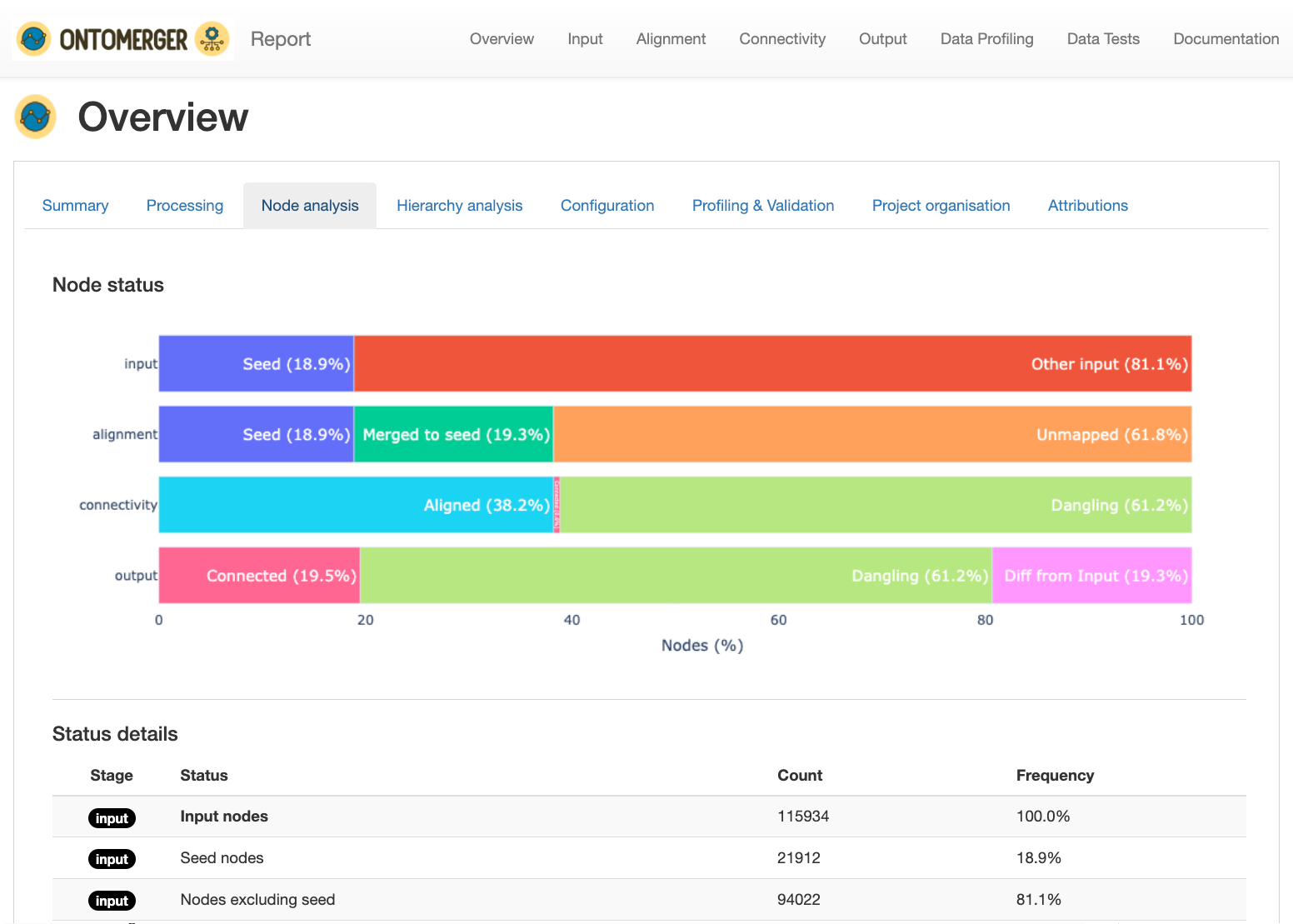}
}
\caption{An example HTML report generated by OntoMerger.} 
\label{fig:report}

\end{figure}  

OntoMerger tackles these challenges by using a scaffolding ontology, together with  mappings and concept hierarchies, to produce an integrated domain ontology with a single DAG (Directed Acyclic Graph) hierarchy. By design, this ontology is not intended to retain complete knowledge for all the integrated concepts (such as hierarchical connections and the granularity of the original conceptualisation), but rather   
relies on additional mappings provided by the users, that are then used to  infer new ones.  
The library is open source\footnote{https://github.com/AstraZeneca/onto\_merger}, implemented in Python~\cite{vanrossum1995python}, and is available to download\footnote{https://pypi.org/project/onto\_merger} as a package via PyPi. 
The library processes data in tabular format using  Pandas~\cite{mckinney2011pandas}.  
OntoMerger produces the hierarchy by converting hierarchy triples into graph objects 
and by running queries for finding paths (using NetworKit~\cite{staudt2016networkit}), as well as for counting 
sub-graphs (using NetworkX~\cite{hagberg2008exploring}).  
To ensure stable performance, input and output data is tested using the 
Great Expectations~\cite{greatexpectations} validation framework.  
In addition, the library provides an in-depth analysis by using the the Pandas Profiling package\footnote{https://github.com/ydataai/pandas-profiling/} to profile all the data that is consumed and produced. Insights about the 
\begin{inparaenum}
\item[(i)] input, 
\item[(ii)] intermediate and produced data (i.e. the domain ontology), 
\item[(iii)] results of data profiling and data validation, and
\item[(iv)]  metrics about the inner workings of the library 
\end{inparaenum}
are presented in an HTML report, shown in Figure \ref{fig:report}.

\section{Related work}\label{sec:related_work}

Ontology integration has been the focus of the research community for several years~\cite{osman2021survey}. Whilst \emph{ontology matching}~\cite{euzenat2007ontology} focuses on the automated alignment, or generation of sets of mappings between the corresponding elements of different ontologies, the use of these mappings to produce a coherent, integrated ontology is non-trivial. This is due to the need to satisfy \emph{merge requirements}~\cite{pottinger2003merging} such as: preserving correctness and coherence; maintaining acyclicity; and avoiding redundancy. To satisfy these requirements, existing algorithms either modify the generated  alignments~\cite{stoilos2018novel,kr2016_negotiation, Solimando2017,udrea2007iliads} or the ontologies themselves~\cite{jimenez-ruiz2009integration,raunich2014atom,fahad2012dkpaom}. 

Despite the significant effort invested in the ontology integration problem, existing methods do not always satisfy the requirements of real-world use cases in the context of large scale data science, which can
compromises their pragmatic use.
For example, existing systems usually rely on OWL semantics and on the explicit constraints defined in source ontologies. However, implicit constraints may also exist; for example, two concepts in the same ontology may not be explicitly declared as disjoint, but this may be implicitly desirable to avoid them being merged as a result of ontology integration.

\section{Preliminaries}
\label{sec:preliminaries}

In this paper, we assume familiarity with basic notions of RDF \cite{klyne2004resource}, ontology alignment and integration \cite{euzenat2007ontology}.

\begin{definition}[Concept Name Set]
\label{def:concept_set}
The symbol $\mathsf{C}$ denotes a set of concept names $\mathsf{C = \{ c_{1}, ...,c_{n} \} }$ where $|\mathsf{C}| = n$. 
A set of concept names from the same ontology is denoted as $\mathsf{C}_{\mathcal{O}}$.
\end{definition}

\begin{definition}[Mapping and Mapping Set]
\label{def:concept_mapping_and_set}
A concept mapping is a triple $\mathsf{m} = \tuple{\mathsf{c}, \mathsf{r}, \mathsf{c'}}$, asserting that some relation $\mathsf{r}$ holds between concepts $\mathsf{c}$ and $\mathsf{c'}$, where 
either $\{ \mathsf{c}, \mathsf{c'} \} \in \mathsf{C}_{\mathcal{O}}$, 
or $\mathsf{c} \in \mathsf{C}_{\mathcal{O}}$ and $\mathsf{c}^\prime \in \mathsf{C}_{\mathcal{O}'}$,
given the ontologies $\mathcal{O}$ and  $\mathcal{O}'$.
The symbol $\mathsf{M}$ denotes a set of concept mappings $\mathsf{M = \{ m_{1}, ...,m_{n} \} }$, $|\mathsf{M}| = n$.
\end{definition}

The semantics of a \textit{mapping relation} $\mathsf{r}$ may vary due to the diversity of input mappings that originate from different sources in real world scenarios.  The may be formally defined~\cite{euzenat2007ontology};  $\mathsf{r \in \{ \equiv, \sqsubseteq, \sqsupseteq, \perp \}}$, or may be loosely defined, such as $\mathsf{r} \in \{ \approxeq, \textsc{db\_xref} \}$, etc. The concepts $\mathsf{c}$ and $\mathsf{c}'$ refer to the \emph{source} and \emph{target} concepts respectively.


\begin{definition}[Hierarchy Edge and Hierarchy Edge Set]
\label{def:concept_hierarchy_edge_and_set}
Given two ontologies,  $\mathcal{O}$ and $\mathcal{O}'$, a concept hierarchy edge is a triple $\mathsf{h} = \tuple{\mathsf{c}, \sqsubseteq, \mathsf{c'}}$, which asserts that a subsumption relation ($\sqsubseteq$) holds between concepts $\mathsf{c}$ and $\mathsf{c'}$, where either $\{ \mathsf{c}, \mathsf{c'} \} \in \mathcal{O}$,  or 
$\mathsf{c} \in \mathcal{O}$ and $\mathsf{c}^\prime \in \mathcal{O}'$.
The symbol $\mathsf{H}$ denotes the concept hierarchy edge set $\mathsf{H = \{ h_{1}, ...,h_{n} \} }$, with $|\mathsf{H}| = n$.
\end{definition}

\begin{definition}[Merge and Merge Set]
\label{def:concept_merge_and_set}
A concept merge is a triple  $\mathsf{x} = \tuple{\mathsf{c}, \Rightarrow, \mathsf{c'}}$, 
asserting that a concept $\mathsf{ c}$ can be replaced by $\mathsf{c}'$ in a concept, mapping or in a hierarchy edge set.
The symbol $\mathsf{X}$ denotes the set of merges $\mathsf{X = \{ x_{1}, ...,x_{n} \} }$, with  $|\mathsf{X}| = n$.
\end{definition}

\begin{definition}[Stable Merge Set]
\label{def:stable_merge_set}
A merge set is $\mathsf{X}$ is stable if  a concept only appears in one merge $\mathsf{x} \in \mathsf{X}$ and only as a source concept.
\end{definition}

In the definitions of mapping, hierarchy edge, and merge triples, we refer to $\mathsf{c}$ as the \textit{source} and $\mathsf{c}'$ as the \textit{target} concept. The function $\textsc{Sig}$ returns the \textit{source} ($\textsc{Sig}_{source}$), \textit{target} ($\textsc{Sig}_{target}$) or both ($\textsc{Sig}$) set of concept names from either an individual mapping or from a set of mappings, merges,  hierarchy edges or paths.

\section{Integration design}\label{sec:model}


The integration design process requires the following \textit{input} sets:
concept names $\mathsf{C}_{in}$, 
obsolete (i.e. deprecated) concept names $\mathsf{C}_{obs}$ 
($\mathsf{C}_{obs} \subseteq \mathsf{C}_{in}$ and $0 \leq |\mathsf{C}_{obs}| \leq |\mathsf{C}_{in}|$),
mappings $\mathsf{M}_{in}$,
and hierarchy edges $\mathsf{H}_{in}$; where each set
may contain members from different ontologies.
In addition, a \textit{configuration} must specify the concept names in a $\textsc{Seed}$ ontology, 
such that $\mathsf{C}_{\mathcal{O}_{\textsc{Seed}}} \subseteq \mathsf{C}_{in}$ and
$\mathsf{H}_{\mathcal{O}_{\textsc{Seed}}} \subseteq \mathsf{H}_{in}$.  
No other ontology must be fully 
contained in the input sets other than the $\textsc{Seed}$ ontology.

The optional \textit{preference order} in which the concepts are de-duplicated and 
connected (denoted as $\textsc{Sources}$) can also be specified.  
If it is omitted, it will be computed by taking the frequency 
of each ontology concept set of $\mathsf{C}_{in}$ in decreasing order.  
In either case, the $\textsc{Seed}$ ontology will be the first item in $\textsc{Sources}$.
The configuration must also contain the $\textsc{MappingGroups}$ list that categorises the mapping types of $\mathsf{M}_{in}$ into a minimum of two categories: \textit{equivalence} and \textit{reference} 
(where the \textit{equivalence} group must have at least one member, and the other groups can be empty sets).  
These categories represent the strength of the mapping relations and are handled in sequence 
(in decreasing order starting from the strongest group) during the deduplication stage.


\begin{algorithm}
\caption{$\textsc{ComputeDomainOntology}$}
\label{alg:alignAndConnectConcepts}
    \DontPrintSemicolon 
    \SetKwInOut{Input}{Input}\SetKwInOut{Intermediate}{Intermediate}\SetKwInOut{Output}{Output} 
	\Input{
	    $\mathsf{C}_{in}$: input concepts;
	    $\mathsf{C}_{obs}$: obsolete concepts of $\mathsf{C}_{in}$;
	    $\mathsf{M}_{in}$: input mappings;
	    $\mathsf{H}_{in}$: input hierarchy edges;
        $\textsc{Seed}$: name of the seed ontology  (i.e.$\mathcal{O}_{\textsc{Seed}}$);
	    $\textsc{Sources}$: preference order;
	    $\textsc{MappingTypes}$: mapping type groups;
	}
	\Intermediate{
	    $\mathsf{C}_{unm}$: unmerged concepts;
    }
	\Output{
	    $\mathsf{C}_{dom}$: domain concepts;
	    $\mathsf{M}_{dom}$: domain mappings;
	    $\mathsf{X}_{can}$: concept merges with canonical target concepts;
	    $\mathsf{H}_{dom}$: domain hierarchy edge set;
    }
    $\mathsf{C}_{dom}, \mathsf{C}_{unm}, \mathsf{X}_{can}, \mathsf{M}_{dom} \leftarrow \textsc{DeduplicateConcepts}(\newline \hspace*{2em} \mathsf{C}_{in}, \mathsf{C}_{obs}, \mathsf{M}_{in},
    \textsc{Seed}, \textsc{Sources}, \textsc{MappingTypes})$ \;
    $\mathsf{H}_{dom} \leftarrow \textsc{ConnectConcepts}(\newline \hspace*{2em} 
    \mathsf{C}_{unm}, \mathsf{X}_{can}, \mathsf{H}_{in}, \textsc{Seed}, \textsc{Sources})$ \;
    \Return $\mathsf{C}_{dom}, \mathsf{M}_{dom}, {X}_{can}, \mathsf{H}_{dom}$ \;     
\end{algorithm}

The main part of \textit{OntoMerger} produces the \textit{domain ontology} by first deduplicating 
concepts (Section \ref{sec:alignment}) and then connecting the remaining 
unmerged concepts (i.e. concepts without an accepted mapping) together to form a single DAG, i.e. concept hierarchy (\ref{sec:connectivity}).  
This process is formalised by Algorithm \ref{alg:alignAndConnectConcepts}.

\subsection{Deduplication}\label{sec:alignment}

\begin{algorithm}
\caption{$\textsc{DeduplicateConcepts}$}
\label{alg:alignConcepts}
    \DontPrintSemicolon 
    \SetKwInOut{Input}{Input}\SetKwInOut{Output}{Output} 
	\Input{
	    $\mathsf{C}_{in}$;
	    $\mathsf{C}_{obs}$;
	    $\mathsf{M}_{in}$;
	    $\textsc{Seed}$;
	    $\textsc{Sources}$;
	    $\textsc{MappingTypes}$
	}
	\Output{
	    $\mathsf{C}_{dom}$; 
	    $\mathsf{C}_{unm}$: unmerged concepts;
	    $\mathsf{X}_{can}$; 
	    $\mathsf{M}_{dom}$; 
    }
    \tcc{Pre-processing}
    $\mathsf{X}_{obs}, \mathsf{M}_{internal} \leftarrow 
        \textsc{ComputeObsoleteMerges}(\mathsf{M}_{in}, \mathsf{C}_{obs})$ \;
    $\mathsf{M}_{in}' \leftarrow \textsc{UpdateMappings}((\mathsf{M}_{in} 
        \setminus \mathsf{M}_{internal}), \mathsf{X}_{obs})$ \;
        
    \tcc{Deduplication}
    $\mathsf{X} \leftarrow \mathsf{X}_{obs} \cup \textsc{ComputeMerges}(\newline \hspace*{2em} 
    \mathsf{C}_{in}, \mathsf{M}_{in}', \textsc{Seed}, \textsc{Sources}, \textsc{MappingTypes})$ \;
        
    \tcc{Post-processing}
    $\mathsf{X}_{can} \leftarrow \textsc{AggregateMerges}(\mathsf{X}, \textsc{Sources})$ \;
    $\mathsf{C}_{dom} \leftarrow \textsc{ApplyMerges}(\mathsf{C}_{in}, \mathsf{X}_{can})$ \;
    $\mathsf{M}_{dom} \leftarrow \textsc{ApplyMerges}(\mathsf{M}_{dom}, \mathsf{X}_{can})$ \;
    $\mathsf{C}_{unm} \leftarrow \mathsf{C}_{in} \setminus 
        (\mathsf{C}_{obs} \cup \mathsf{Sig}_{source}(\mathsf{X}_{can}))$ \;
    \Return $\mathsf{C}_{dom}, \mathsf{C}_{unm}, {X}_{can}, \mathsf{M}_{dom}$ \;     
\end{algorithm}

The deduplication stage, formalised by Algorithm \ref{alg:alignConcepts}, reduces the
input set of concepts ($\mathsf{C}_{in}$) by processing mappings to create a \textit{stable} concept merge set ($\mathsf{X}_{can}$). 

\subsubsection{Pre-processing.} First, we compute the merges for obsolete concepts.
Some mappings assert a relation between concepts of the same ontology to describe internal 
concept renaming, i.e. when a concept is deprecated and assigned to a new name.  
Using the set of obsolete concepts, the function $\textsc{ComputeObsoleteMerges}$ computes 
merges $\mathsf{X}_{obs}$ from the relevant input mappings 
$\{ \mathsf{m} = \tuple{\mathsf{c}, \equiv, \mathsf{c'}} \in \mathsf{M}_{in}  |  \{ \mathsf{c}, \mathsf{c'} \} \in \mathcal{O}, \mathsf{r} = \tuple{\mathsf{c}, \Rightarrow, \mathsf{c'}}$ where $\mathsf{c} \in \mathsf{C}_{obs} \wedge \mathsf{c'} \not \in \mathsf{C}_{obs}\}$.  
Next, we update the mappings by applying the obsolete concept merges $\mathsf{X}_{obs}$, so the output $\mathsf{M}_{in}'$ now only contains current concept names.

\begin{algorithm}
\caption{$\textsc{ComputeMerges}$}
\label{alg:compMerges}
    \DontPrintSemicolon 
    \SetKwInOut{Input}{Input}\SetKwInOut{Output}{Output} 
	\Input{
	    $\mathsf{C}_{in}$; 
	    $\mathsf{M}_{updated}$: mappings with current concept names;
	    $\mathsf{X}_{obs}$;
	    $\textsc{Seed}$;
	    $\textsc{Sources}$;
	    $\textsc{MappingTypes}$;
    }
	\Output{
	    $\mathsf{X}$: concept merges
    }
    $\mathsf{C}_{unm} \leftarrow \mathsf{C}_{in} \setminus \textsc{GetConcepts}(\mathsf{C}_{in}, \textsc{Seed})$ \;
    $\mathsf{X} \leftarrow \emptyset$ \;
    \For{each $\textsc{M\_Type} \in \textsc{MappingTypes}$} {
        \For{each $\textsc{Source} \in \textsc{Sources}$} {
            $\mathsf{M}_{Source} \leftarrow \textsc{GetMps}(\mathsf{M}_{updated}, \textsc{M\_Type}, \textsc{Source})$ \;
            $\mathsf{M}_{Source}' \leftarrow \textsc{Filter}(\mathsf{M}_{Source}, \mathsf{C}_{unm})$ \;
            $\mathsf{M}_{Source}'' \leftarrow \textsc{OrientTo}(\mathsf{M}_{Source}', \textsc{Source})$ \;
            $\mathsf{M}_{Source}''' \leftarrow \textsc{Get1Or*SourceTo1Target}(\mathsf{M}_{Source}'')$ \;
            $\mathsf{X} \leftarrow \mathsf{X} \cup \textsc{ConvertToMerges}(\mathsf{M}_{Source}''')$ \;
            $\mathsf{C}_{unm} \leftarrow \mathsf{C}_{unm} \setminus \mathsf{Sig}_{source}(\mathsf{X}) \;$
        }
    }
    \Return ${X}$ \;  
\end{algorithm}

\subsubsection{Alignment.} Algorithm \ref{alg:compMerges} computes the stable merge set.  
The set of concepts $\mathsf{C}_{unm}$ 
keeps track of unmerged  (i.e. not merged) concepts; this initially contains all input concepts, excluding concepts of the seed ontology.  
The alignment process loops through the mapping type groups and the ontology alignment order. In each iteration, we attempt to align every remaining unmerged concept to concepts of the 
given $\textsc{Source} \in \textsc{Sources}$.  The function $\textsc{GetMps}$ returns all 
mappings that are in the specified type group $\textsc{M\_Type}$ and map concepts of the given $\textsc{Source}$; this ($\mathsf{M}_{Source}$) is filtered by the function $\textsc{Filter}$, so the resulting mapping set $\mathsf{M}_{Source}'$ only contains those mappings that align unmerged concepts.  
These mappings focus on the current $\textsc{Source}$ such that the target of mappings the is always from $\mathsf{M}_{Source}$.  Finally, the oriented mapping set $\mathsf{M}_{Source}''$ is further filtered by multiplicity with the function $\textsc{Get1Or*SourceTo1Target}$ such that in the produced mapping set ($\mathsf{M}_{Source}'''$) every source concept is mapped to exactly one target concept (but one target may be mapped to many source concepts).  
The mappings are then converted to merges and stored in $\mathsf{X}$, and the mapped concepts are removed from $\mathsf{C}_{unm}$.

\subsubsection{Post-processing.} The function $\textsc{AggregateMerges}$ groups the merge set $\mathsf{X}$ and rewrites them such that the target concept is always the canonical concept name of the given \textit{merge cluster}.  Concepts of each merge group form a cluster, where only one of these 
concepts is the canonical concept; this is determined by the preference order ($\textsc{Sources}$).  
For example, given the merge group $\{ \tuple{\mathsf{S_{1}:001}, \Rightarrow, \mathsf{S_{2}:001}}, \tuple{\mathsf{S_{2}:001}, \Rightarrow, \mathsf{S_{3}:001}} \}$ and the priority order $(\mathsf{S_{3}}, \mathsf{S_{2}}, \mathsf{S_{1}})$, $\{ \tuple{\mathsf{S_{1}:001}, \Rightarrow, \mathsf{S_{3}:001}}, \tuple{\mathsf{S_{2}:001}, \Rightarrow, \mathsf{S_{3}:001}} \}$ is the canonical merge set, where $\mathsf{S_{3}:001}$ is the canonical concept.  
Finally, we compute the domain concept (${C}_{dom}$) and mapping sets (${M}_{dom}$) by applying the canonical merges (${X}_{can}$) to replace each concept name with the corresponding canonical concept (using function $\textsc{ApplyMerges}$), and compute the set of final unmerged concepts (${C}_{unm}$).

\subsection{Connectivity}\label{sec:connectivity}

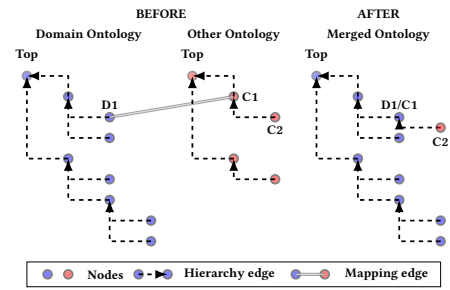
\begin{figure}[h!]
\vspace{-0.3cm}
\centering
\scalebox{0.55}{
\begin{tikzpicture}

\node[circle,draw=black, fill=blue, inner sep=0pt,minimum size=6pt, opacity=0.4, very thick] (b) at (-3,4) {};

\node[circle,draw=black, fill=blue, inner sep=0pt,minimum size=6pt, opacity=0.5, very thick] (b) at (0,0) {};
\node[circle,draw=black, fill=blue, inner sep=0pt,minimum size=6pt, opacity=0.5, very thick] (b) at (0,0.5) {};
\node[circle,draw=black, fill=blue, inner sep=0pt,minimum size=6pt, opacity=0.5, very thick] (b) at (-1,1) {};

\node[circle,draw=black, fill=blue, inner sep=0pt,minimum size=6pt, opacity=0.5, very thick] (b) at (-1,1.5) {};
\node[circle,draw=black, fill=blue, inner sep=0pt,minimum size=6pt, opacity=0.5, very thick] (b) at (-2,2) {};

\node[circle,draw=black, fill=blue, inner sep=0pt,minimum size=6pt, opacity=0.5, very thick] (b) at (-1,2.5) {};

\node[circle,draw=black, fill=blue, inner sep=0pt,minimum size=6pt, opacity=0.5, very thick] (b) at (-1,3) {};

\node[circle,draw=black, fill=blue, inner sep=0pt,minimum size=6pt, opacity=0.5, very thick] (b) at (-2,3.5) {};

\draw[black, dashed, very thick] (0,0.5) -- (-1,0.5);

\draw[black, dashed, very thick] (-1,1.5) -- (-2,1.5);

\draw[black, dashed, very thick] (-1,3) -- (-2,3);

\draw[black, dashed, very thick, -Latex] (0,0) -- (-1,0) -- (-1,1);

\draw[black, dashed, very thick, -Latex] (-1,1) -- (-2,1) -- (-2,2);

\draw[black, dashed, very thick, -Latex] (-2,2) -- (-3,2) -- (-3,4);

\draw[black, dashed, very thick, -Latex] (-1,2.5) -- (-2,2.5) -- (-2,3.5);

\draw[black, dashed, very thick, -Latex] (-2,3.5) -- (-2,4) -- (-3,4);


\node[circle,draw=black, fill=red, inner sep=0pt,minimum size=6pt, opacity=0.4, very thick] (b) at (1,4) {};

\node[circle,draw=black, fill=red, inner sep=0pt,minimum size=6pt, opacity=0.5, very thick] (b) at (3,1.5) {};
\node[circle,draw=black, fill=red, inner sep=0pt,minimum size=6pt, opacity=0.5, very thick] (b) at (2,2) {};

\node[circle,draw=black, fill=red, inner sep=0pt,minimum size=6pt, opacity=0.5, very thick] (b) at (3,3) {};

\node[circle,draw=black, fill=red, inner sep=0pt,minimum size=6pt, opacity=0.5, very thick] (b) at (2,3.5) {};

\draw[black, dashed, very thick, -Latex] (3,1.5) -- (2,1.5) -- (2,2);

\draw[black, dashed, very thick, -Latex] (3,3) -- (2,3)-- (2,3.5);

\draw[gray, opacity=0.7, thick, double] (-1,3) -- (2,3.5);

\draw[black, dashed, very thick, -Latex] (2,2) -- (1,2) -- (1,4);

\draw[black, dashed, very thick, -Latex] (2,3.5) -- (2,4) -- (1,4);


\node[circle,draw=black, fill=blue, inner sep=0pt,minimum size=6pt, opacity=0.4, very thick] (b) at (4,4) {};

\node[circle,draw=black, fill=blue, inner sep=0pt,minimum size=6pt, opacity=0.5, very thick] (b) at (7,0) {};
\node[circle,draw=black, fill=blue, inner sep=0pt,minimum size=6pt, opacity=0.5, very thick] (b) at (7,0.5) {};
\node[circle,draw=black, fill=blue, inner sep=0pt,minimum size=6pt, opacity=0.5, very thick] (b) at (6,1) {};

\node[circle,draw=black, fill=blue, inner sep=0pt,minimum size=6pt, opacity=0.5, very thick] (b) at (6,1.5) {};
\node[circle,draw=black, fill=blue, inner sep=0pt,minimum size=6pt, opacity=0.5, very thick] (b) at (5,2) {};

\node[circle,draw=black, fill=blue, inner sep=0pt,minimum size=6pt, opacity=0.5, very thick] (b) at (6,2.5) {};

\node[circle,draw=black, fill=blue, inner sep=0pt,minimum size=6pt, opacity=0.5, very thick] (b) at (6,3) {};

\node[circle,draw=black, fill=blue, inner sep=0pt,minimum size=6pt, opacity=0.5, very thick] (b) at (5,3.5) {};

\draw[black, dashed, very thick] (7,0.5) -- (6,0.5);

\draw[black, dashed, very thick] (6,1.5) -- (5,1.5);

\draw[black, dashed, very thick] (6,3) -- (5,3);

\draw[black, dashed, very thick, -Latex] (7,0) -- (6,0) -- (6,1);

\draw[black, dashed, very thick, -Latex] (6,1) -- (5,1) -- (5,2);

\draw[black, dashed, very thick, -Latex] (5,2) -- (4,2) -- (4,4);

\draw[black, dashed, very thick, -Latex] (6,2.5) -- (5,2.5) -- (5,3.5);

\draw[black, dashed, very thick, -Latex] (5,3.5) -- (5,4) -- (4,4);

\draw[black, dashed, very thick, -Latex] (7,2.75) -- (6,2.75) -- (6,3);

\node[circle,draw=black, fill=red, inner sep=0pt,minimum size=6pt, opacity=0.5, very thick] (b) at (7,2.75) {};

\node at (0.25,5.5) {\textbf{BEFORE}};

\node at (-1.5,5) {\textbf{Domain Ontology}};

\node at (2,5) {\textbf{Other Ontology}};

\node at (5.5,5.5) {\textbf{AFTER}};

\node at (5.5,5) {\textbf{Merged Ontology}};

\node at (-3,4.5) {\textbf{Top}};

\node at (1,4.5) {\textbf{Top}};

\node at (4,4.5) {\textbf{Top}};

\node at (-1,3.3) {\textbf{D1}};

\node at (2.4,3.5) {\textbf{C1}};

\node at (3,2.7) {\textbf{C2}};

\node at (6,3.3) {\textbf{D1/C1}};

\node at (7,2.4) {\textbf{C2}};

\draw[black] (7,-0.5) -- (-3,-0.5) -- (-3,-1.1) -- (7, -1.1) -- (7, -0.5);

\node[circle,draw=black, fill=blue, inner sep=0pt,minimum size=6pt, opacity=0.5, very thick] (b) at (-2.5,-0.8) {};

\node[circle,draw=black, fill=red, inner sep=0pt,minimum size=6pt, opacity=0.5, very thick] (b) at (-2,-0.8) {};

\node at (-1.1,-0.8) {\textbf{Nodes}};

\node[circle,draw=black, fill=blue, inner sep=0pt,minimum size=6pt, opacity=0.5, very thick] (b) at (-0.3,-0.8) {};

\node[circle,draw=black, fill=blue, inner sep=0pt,minimum size=6pt, opacity=0.5, very thick] (b) at (0.4,-0.8) {};

\draw[black, dashed, very thick, -Latex] (-0.3,-0.8)  -- (0.4,-0.8);

\node at (1.9,-0.8) {\textbf{Hierarchy edge}};

\node[circle,draw=black, fill=blue, inner sep=0pt,minimum size=6pt, opacity=0.5, very thick] (b) at (3.5,-0.8) {};

\node[circle,draw=black, fill=red, inner sep=0pt,minimum size=6pt, opacity=0.5, very thick] (b) at (4.2,-0.8) {};

\node at (5.7,-0.8) {\textbf{Mapping edge}};

\draw[gray, opacity=0.7, thick, double] (3.5,-0.8) -- (4.2, -0.8);

\end{tikzpicture}}
\caption{Integrating an unmapped concept via its original hierarchy path and a mapped parent.}
\vspace{-0.45cm}
\end{figure}

\begin{algorithm}
\caption{\textsc{ConnectConcepts}}
\label{alg:connectConcepts}
    \DontPrintSemicolon 
    \SetKwInOut{Input}{Input}\SetKwInOut{Output}{Output} 
	\Input{
	    $\mathsf{C}_{unm}$; 
	    $\mathsf{H}_{in}$;
	    $\mathsf{X}_{can}$; 
	    $\textsc{Seed}$;
	    $\textsc{Sources}$;
	}
	\Output{
	    $\mathsf{H}_{dom}$: output hierarchy edge set;
    }
    $\mathsf{H}_{dom} \leftarrow \textsc{GetHierarchy}(\textsc{Seed}, \mathsf{H}_{in})$ \;
    $\mathsf{C}_{connected} \leftarrow \textsc{Sig}(\mathsf{H}_{dom}) \cup 
        \textsc{Sig}_{source}(\textsc{GetMergesToSource}(\textsc{Seed}, \mathsf{X}_{can})) $ \;
    \For{each $\textsc{Source} \in \textsc{Sources}$} {
        $\mathsf{H}_{\textsc{Source}} \leftarrow \textsc{GetHierarchy}(\textsc{Source}, \mathsf{H}_{in})$ \;
        $\mathsf{C}_{\textsc{UnmappedSource}} \leftarrow \textsc{GetConcepts}(\mathsf{C}_{unm}, \textsc{Source})$ \;
        \For{each $\mathsf{c} \in \mathsf{C}_{\textsc{UnmappedSource}}$} {
            $\mathsf{Path}_{\mathsf{c}} \leftarrow \textsc{GetShortestPathToRoot}(\mathsf{c}, \mathsf{H}_{\textsc{Source}})$ \;
            \If{$ \textsc{Sig}(\mathsf{Path}_{\mathsf{c}}) \cap \mathsf{C}_{connected} \not = \emptyset$} {
                $\mathsf{Path}_{\mathsf{c}}' \leftarrow \textsc{Prune}(\mathsf{Path}_{\mathsf{c}}, \mathsf{C}_{connected})$ \;
                $\mathsf{H}_{\mathsf{c}} \leftarrow \textsc{ConvertToEdges}(\mathsf{Path}_{\mathsf{c}})$ \;
                $\mathsf{C}_{connected} \leftarrow \mathsf{C}_{connected} \cup \textsc{Sig}(\mathsf{H}_{\mathsf{c}})$ \;
                $\mathsf{H}_{dom} \leftarrow \mathsf{H}_{dom} \cup \mathsf{H}_{\mathsf{c}}$ \;
            }
        }
    }
    \Return $\mathsf{H}_{domain}$ \;     
\end{algorithm} 

After concept deduplication, the remaining concepts ($\mathsf{H}_{unm}$) are organised into a DAG, i.e. a concept hierarchy using the input hierarchy edge set ($\mathsf{H}_{in}$).  This process is formalised by Algorithm \ref{alg:connectConcepts}, and the main idea, which is a simplified version of the approach presented in~\cite{stoilos2018novel,juric2021platform}, is depicted in Figure 1. 
First, we add the seed ontology hierarchy to the domain hierarchy set ($\mathsf{H}_{dom}$) and initialise the set of connected nodes with the seed hierarchy concepts and the concepts that were merged to seed concepts.  
Then we iterate through each source in the priority order and filter out the corresponding hierarchy fragment from the input set with the function $\textsc{GetHierarchy}(\textsc{Source}, \mathsf{H}_{in})$ (because path finding performs better on smaller inputs).  Next, for each unmerged concept in $\mathsf{H}_{\textsc{Source}}$, we check whether there is a path to the root concept that contains a connected concept (i.e. an anchor point that is either a seed concept, is merged to a seed concept, or has already been connected to $\mathsf{H}_{dom}$).  If there is, we prune the path by dropping those intermediate concepts that are not in $\mathsf{C}_{\textsc{UnmappedSource}}$ and convert the pruned path to a set of hierarchy edges.  Finally, we update $\mathsf{C}_{connected}$ and $\mathsf{H}_{dom}$ before proceeding to the next unmerged concept.  The result is a crude approximation where we may lose concept classification granularity (i.e. the dropped intermediate concepts) in order to establish a connection that does not increase the size of the domain concept set but maximises the hierarchy coverage of $\mathsf{C}_{unm}$ by  $\mathsf{H}_{dom}$.

\section{Experimental Evaluation}\label{sec:evaluation}


We conducted an experimental evaluation on a \textit{real-life data set}: the \textit{disease}  
concepts of BIKG \cite{geleta2021biological} in order to assess the effectiveness 
of our approach in reducing duplication and creating connectivity.  
A comparative evaluation was not possible since currently there are no other systems that 
operate with the same requirements as OntoMerger~\footnote{one of the main requirements being availability in Python}.  Here we only include a brief evaluation; however, the library produces an in-depth 
analysis report~\footnote{\url{https://htmlpreview.github.io/?https://github.com/AstraZeneca/onto_merger/blob/main/data/example_report/index.html}}, as depicted in Figure \ref{fig:report}, using several  metrics~\cite{stoilos2018methods} to evaluate the KG integration.
\begin{table}[!ht]
\centering
\footnotesize
    \begin{tabular}{lrr}
    \hline
        \textbf{Metric} & \textbf{Count (Input)} & \textbf{Count (Output)} \\ \hline 
        Concept sources (ontologies) & 16 & 16 \\  
        Concepts & 187,717 & 152,021 \\ 
        Concept merges & n/a & 35,696 \\ 
        Mappings & 439,389 & 133,496  \\ 
        Connected subgraphs & 4 & 1  \\ 
        Hierarchy edges & 272,824 & 60,621 \\   \hline
    \end{tabular}

\caption{Input and output (domain ontology) comparison}
\label{tab:input_output_comparison}
\vspace{-0.5cm}
\end{table}
Table \ref{tab:input_output_comparison} 
shows a comparison of the input and output data, while Table \ref{tab:node_status_deduplication_and_connectivity} 
presents the results of the deduplication of concepts and the creation of the concept 
hierarchy.  30.79\% of input concepts (of 16 sources) were merged, 
and 41.11\% were successfully integrated into a DAG using only 4 hierarchies.
The overall run time of the process was 25 minutes on an 
average machine (6 core CPU, 32GB memory).  The majority of the run time was
the generation of the concept hierarchy (72\%) and the result analysis (13\%).

\begin{table}[!ht]
\centering
\footnotesize
    \begin{tabular}{clrr}
    \hline
        \multirow{2}{*}{\textbf{Stage}} & \textbf{Concept set} & \textbf{Count} & \textbf{Ratio (\%)} \\ \cline{2-4}
         & \textit{Input} & 187,717 & 100.00\% \\ \hline 
        \multirow{4}{*}{Deduplication} & (Mapped) Seed & 21,910 & 11.78\% \\ 
         & (Mapped) to Seed & 34,333 & 18.28\% \\ 
         & (Mapped) to other (excluding to seed) & 1,363 & 0.72\% \\
         & \textit{Unmapped} & 129,912 & 69.21\% \\ \hline
         \multirow{3}{*}{Connectivity} & (Connected) Seed + Mapped to Seed & 56,442 & 30.06\% \\ 
         & (Connected) & 20,740 & 11.05\% \\ 
         & \textit{Disconnected} & 109,172 & 58.16\% \\ \hline
    \end{tabular}
\caption{Concept deduplication and connectivity status}
\label{tab:node_status_deduplication_and_connectivity}
\vspace{-0.92cm}
\end{table}

\section{Conclusions}

In this paper we have presented OntoMerger, an ontology integration library that deduplicates KG concepts of the same domain, and organises them into a single DAG, i.e. a concept hierarchy.  
Furthermore, we have empirically evaluated the approaches that perform deduplication of concepts and the formation of the concept hierarchy.  In this section, we address the limitations of the library and outline directions for our future work.

\textit{Limitations.} The library does not directly consume traditional RDF formats but requires the additional step of parsing and transforming concept names, mappings, and hierarchy edges into a columnar data structure.  Moreover, the output is also in a columnar format.
The library handles all data in-memory, posing several limitations regarding the scalability of the system.  
Firstly, processing tables with Pandas becomes challenging for large datasets.  
Secondly, hierarchy path computation slows down significantly due to using in-memory graphs.
Currently, only the shortest path is computed for each concept when we attempt to connect 
them to the domain hierarchy.  However, in many ontologies, there are often multiple possible 
paths to the root of the ontology.  Using only one path could lead to missing out on paths that contain an anchor point, i.e. a connected concept.

\textit{Future Directions.} In order to get a better view of the effectiveness of this approach, we plan to conduct a comparative evaluation of our system with several available (Java) ontology integration libraries.
The library could use a mediator step to consume RDF formats directly and translate them into the internal columnar data structures.  There are several libraries available that perform RDF to table translation, for instance KGTK~\cite{ilievski2020kgtk}.

To address scalability when handling large tables, the system could use distributed data processing such as Spark \cite{spark}.  Path computation could be scaled by using an external triple store, where path computation is optimised (this is often achieved by the full materialisation of hierarchy triples)~\cite{10.1007/978-3-319-24369-6_5}.  
Offloading path computation would also enable us to use more than one path during the hierarchy formation process.  However, this could also be improved by either implementing a greedy strategy (keep looking for paths until a suitable one is found) or providing the ability to specify the path computation strategy in the OntoMerger configuration file.
Finally, OntoMerger may merge concepts of the same ontology when it may be more appropriate to convert these merges into hierarchy edges.

\bibliographystyle{ACM-Reference-Format}
\bibliography{main}

\end{document}